\documentclass[journal]{journal}

\usepackage[noadjust]{cite}

\usepackage[inkscapelatex=false]{svg}

\usepackage{subfigure}

\usepackage{subcaption}

\usepackage{graphicx}

\usepackage{booktabs}

\usepackage{longtable}

\usepackage{makecell}

\usepackage{tipa}

\usepackage{footmisc}

\begin{document}

\title{Measurement of the Granularity of Vowel Production Space By Just Producible Difference (JPD) Limens}

\author{Peter Viechnicki}
\thanks{Human Language Technology Center of Excellence, Johns Hopkins University}

\maketitle
\thispagestyle{empty}

\begin{abstract}
A body of work over the past several decades has demonstrated that the complex and coordinated articulatory movements of human vowel production are governed (at least in part) by control mechanisms whose targets are regions of auditory space. Within the target region control at the sub-phonemic level has also been demonstrated.  But the degree of accuracy of that control is unknown.    The current work investigates this question by asking how far apart must two vowel stimuli lie in auditory space in order to yield reliably different imitations? This distance is termed ‘Just Producible Difference’ (JPD). The current study uses a vowel mimicry paradigm to derive the first measurement of JPD among two sets of English speakers during front vowel production.  JPD is estimated at between 14 and 51 mels in $F1 \times F2$ space.

This finding has implications for episodic theories of speech production. It also clarifies the possible structures of human vowel systems, by setting a theoretical lower bound for how close two vowel phonemes may be in a speaker’s formant space, and hence a psychophysical explanation of observed trends in number and patterns of possible vowel phonemes.
\end{abstract}


\IEEEpeerreviewmaketitle

\section{Introduction and Related Work}

\IEEEPARstart{B}{ecause} the motor actions and acoustic outputs of human speech are so complex, scientists have for a long time sought to understand how the speech production system is regulated and coordinated. The nature of speech production targets as combined time-varying trajectories in auditory and somato-sensory space has been well established by decades of experimental findings \cite{perkell_2013}. This research investigates the accuracy of those targets \textemdash termed `granularity' \textemdash for vowel production under mimicry. We first review some salient findings on vowel production targets before describing the current study.

For vowels, the acoustic component of the production target has been shown to be primary over other task variable representations \cite{feng_gracco_max_2011}. Various factors have been posited to influence the locations and overall distribution in the vowel space of the target regions. Among the factors that have been proposed are: 
\begin{itemize}
\item A tendency towards dispersion of vowels within the auditory space \cite{schwartz_etal_1997}; \item Quantal effects – i.e. regions of the vowel space where relatively large articulatory changes yield little acoustic change \cite{STEVENS_1989}; 
\item The balance between communicative efficacy and ease of production \cite{PERKELL2000233}. 
\end {itemize}
Cross-linguistic surveys of extant vowel systems have been used to argue for the influence of each factor on observed distributions \cite{LADEFOGED199093,boe_2002}; this remains an active area of research \cite{yang_2024}.

Multiple lines of research have investigated vowel production targets within broader theories of speech motor control. Vowel production targets are affected by low-latency feedback, as well as longer-latency feed-forward control \cite{PERKELL_2012}. A large body of work has used altered auditory feedback of various kinds to study the properties of vowel targets. Miller et al. 2023 summarizes 22 such studies \cite{miller2023not}, leading to the conclusion that vowel production targets are somewhat plastic at various time scales. They also function interdependently: it has been shown that speakers use knowledge of the entire vowel space to plan their productions \cite{fox2017reconceptualizing}, and adaptation in response to altered feedback are applied to other vowels within the space in varying degrees \cite{cai2010adaptive}. 

It has long been known that speech production targets are influenced by speakers’ perceptual abilities (e.g. \cite{bradlow1997training,BEDDOR2024101352} \textit{inter alia}). Perkell (2012) shows that subjects with more perceptual acuity in the production of a vowel contrast evince less variable productions of the same contrast \cite{PERKELL_2012}.  Such methods lead to static estimates of the distance between vowel production targets for American English (AE) central vowels of between 50 and 100 Hz \cite{perkell2004distinctness}, with the prediction that this distance would vary depending on perceptual acuity, and training. 
Sub-phonemic control of vowel production has been convincingly demonstrated \cite{chistovich_1966,viechnicki_2002}. Recent studies using vowel shadowing show additional evidence for sub-phonemic control of vowel production \cite{tilsen2009subphonemic}, with vowel shifts in F1 of between 2 and 26 Hz and F2 of between 1 and 60 Hz, in response to centralized variants of the vowels /\textipa{I}/ and /a/ with formants shifted by 50 Hz (F1) and 70 Hz (F2).

So far we have been discussing vowel targets as individual goals of speakers considered in isolation.  However, interactional factors have a strong effect on production targets. Many aspects of speech show accommodation to an interlocutor \cite{gessinger2021phonetic}; vowel formants are one aspect where accommodation can be readily demonstrated (e.g. \cite{babel2009selective}).

Vowel mimicry is a useful non-invasive means of investigating the response properties of the speech production system to information provided during conversations. While not natural \textit{per se,} mimicry shares mechanisms with naturally occurring accommodation behaviors \cite{dufour2013much}. Careful collection of perceptual and natural production data from mimicry subjects has been important to disentangling categorical effects from inter-subject behavioral differences \cite{kent1973imitation,repp1987categorical,schouten1977imitation}.  To date, variability in mimicry output has been most successfully explained by stimulus familiarity \cite{nye_fowler_2003}, and by the linguistic ‘meaningfulness’ of the stimulus variant dimensions \cite{kent1979imitation}. 

Exemplar theories of speech motor control (e.g.  \cite{pierrehumbert2002word,pierrehumbert2003exemplar,Johnson2005DecisionsAM}) can model the mechanism by which categorical targets are adjusted in near-real time based on auditory traces held in episodic memory. The characteristics of these episodic auditory representations are not fully clear \cite{cho_feldman_2016}: in particular we do not know the limit of how small or closely packed those adjusted targets can be while still yielding a differential output. This is what we refer to as the ‘granularity’ of the vowel production space.  

\begin{figure}
    \begin{centering}
    \includegraphics[width=\columnwidth]{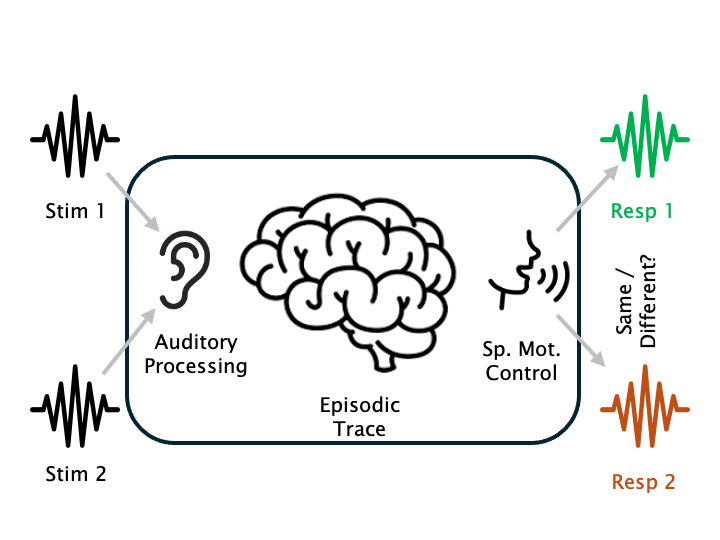}
    \caption{Vowel Mimicry Transfer Function}
    \label{Fig: Mimicry Transfer Function}
    \end{centering}
\end{figure}

More formally, following Chistovich \cite{chistovich_1966} we conceive of the speech production system under mimicry as a function which maps an auditory-perceptual input ($S$, the sound to be imitated) to an acoustic output ($R$, the response sound produced by the mimicker). Given two input stimuli $s_1, s_2$, we then study whether the two mimicked productions $r_1, r_2$ are the same or different to some threshold $t$: $diff(r_1, r_2) > t$?. Our formalization of vowel mimicry differential control is schematized in Fig. \ref{Fig: Mimicry Transfer Function}.

The current study uses a modified vowel mimicry paradigm to investigate the responsiveness of the filter shown in Fig. 1, i.e. how close they could theoretically lie in the vowel space while still leading to differential responses.  The remainder of this article describes the vowel mimicry paradigm used in the two experiments to measure JPD. We then present results and discuss their implications for our understanding of vowel systems. Finally, we describe limitations of the current study and highlight directions for future research. 

\section{Methodology}
\subsection{Overview}
\subsubsection{Goals}
Two vowel mimicry experiments of synthetic vowel continua were carried out, both using interstimulus step sizes small enough to resolve within-category differences \cite{schouten1977imitation}. In addition to mimicry production data, individual perception and natural production data were also collected, to clarify the properties of each subject’s vowel mimicry transfer function. Both experiments elicited mimicry productions of synthetic vowels which were analyzed to yield measurements of the Just Producible Difference (JPD) limen between paired stimuli. The second experiment yielded only partial results due to problems with synthetic stimulus creation.

\subsubsection{Apparatus}
Experimental instructions and stimuli were controlled by custom software running on a Sun Solaris workstation. Recordings were made using a Shure SM96 Condenser 200 microphone connected to a Rane MS1pre-amp and recorded to a Tascam DA-20 MKII Digital Audio Tape. Stimuli were presented to the subjects over Sennheiser HD580 Precision headphones.

\subsubsection{Subjects}
Subject for both experiments were University of Chicago undergraduates or staff, with no reported speech or hearing problems, and native speakers of American English, based on the criterion of having attended elementary school in the United States. Sixteen subjects participated in Experiment 1 (eight male and eight female). Eight subjects (four male, four female) participated in Experiment 2.

\subsection{Experiment 1}
\subsubsection{Stimuli}
\begin{figure}
    \begin{centering}
    \includegraphics[width=\columnwidth]{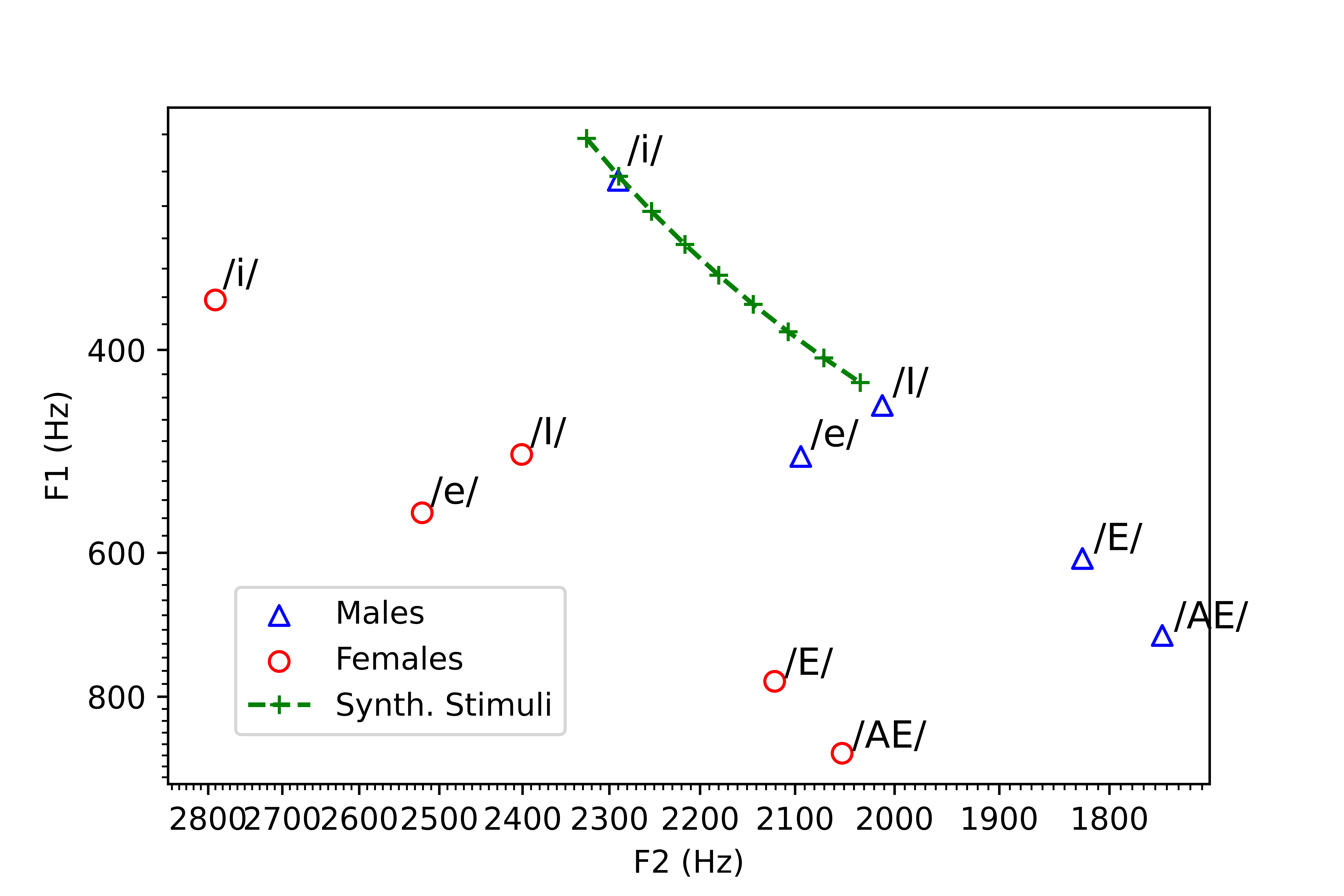}
    \caption{Natural Productions and Synthetic Stimuli}
    \label{Fig: Exp1 Nat Prod}
    \end{centering}
\end{figure}

Nine synthetic stimuli were prepared, a continuum between /i/ and /\textipa{I}/ varying in F1 and F2 only with a constant duration of 250 ms. The endpoints of the continuum were chosen to approximate points slightly beyond the prototypical AE /i/ and /\textipa{I}/ values in $F1 \times F2$ space \cite{peterson_barney_1952}. The seven intermediate tokens were synthesized at equally-space mel steps from the endpoints. Fundamental frequency of all tokens was synthesized with a rise-fall contour for maximum naturalness, with a mean f0 of 117 Hz, giving a male-like voice.  Experiment 1 stimuli in relation to male and female subjects’ natural productions are plotted in $F1 \times F2$ space in Fig. \ref{Fig: Exp1 Nat Prod}.

\subsubsection{Procedure}
Experiment 1 had three components: (1) baseline recordings of natural productions of all AE monophthongs; (2) perceptual categorization and goodness testing of stimulus series; and (3) mimicry elicitation.
Subjects first recorded the 11 contrastive monophthongs of AE four times each in random order by speaking them in the sentence frame ‘Say the word hVd again.’
Next the perceptual categorization and mimicry portions of the experiment were performed. Half the subjects completed the perceptual testing first and then mimicked, while the other half mimicked first and then made perceptual judgements. This manipulation was intended as a control for familiarity effects in mimicry. (No significant effect of experimental order was found in subsequent analyses, so this manipulation was dropped from further consideration.) In the perceptual test, subjects heard each of the nine stimuli six times in random order, and were asked to identify the stimulus using a forced-choice test as /i/ or /\textipa{I}/, and to rate its quality on a 3-point scale. In the mimicry portion, subjects heard each stimulus six times in random order, and were asked to mimic the sound exactly as they heard it. 

\subsubsection{Analysis}\label{Sec: Exp1 Analysis}
\begin{figure}
    \begin{centering}
    \includegraphics[width=\columnwidth]{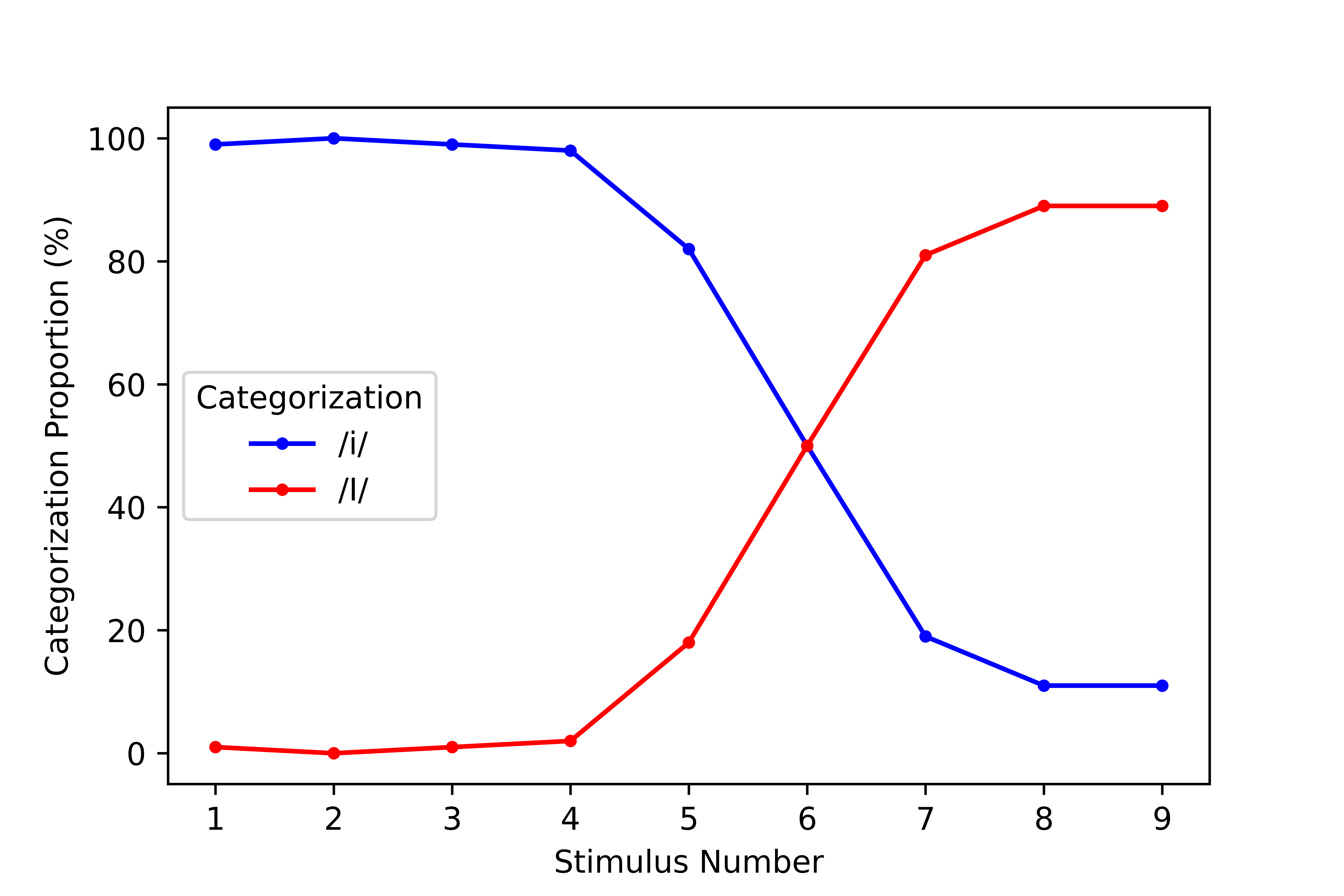}
    \caption{Perceptual Categorization of Exp 1 Stimuli}
    \label{Fig: Exp1 Categorization}
    \end{centering}
\end{figure}

All subjects in post-test interviews reported the stimuli sounded speech-like. Subjects readily categorized the stimuli as /i/ or /\textipa{I}/: categorization curves are shown in Fig. \ref{Fig: Exp1 Categorization} aggregated across all sixteen subjects. Categorization responses were slightly less consistent for the /\textipa{I}/-like stimuli (higher numbers). A larger portion of the stimuli were categorized as /\textipa{I}/, which may be due to phonotactic constraints of AE which do not license /\textipa{I}/ in open syllables.

\begin{figure}
    \begin{centering}
    \includegraphics[width=\columnwidth]{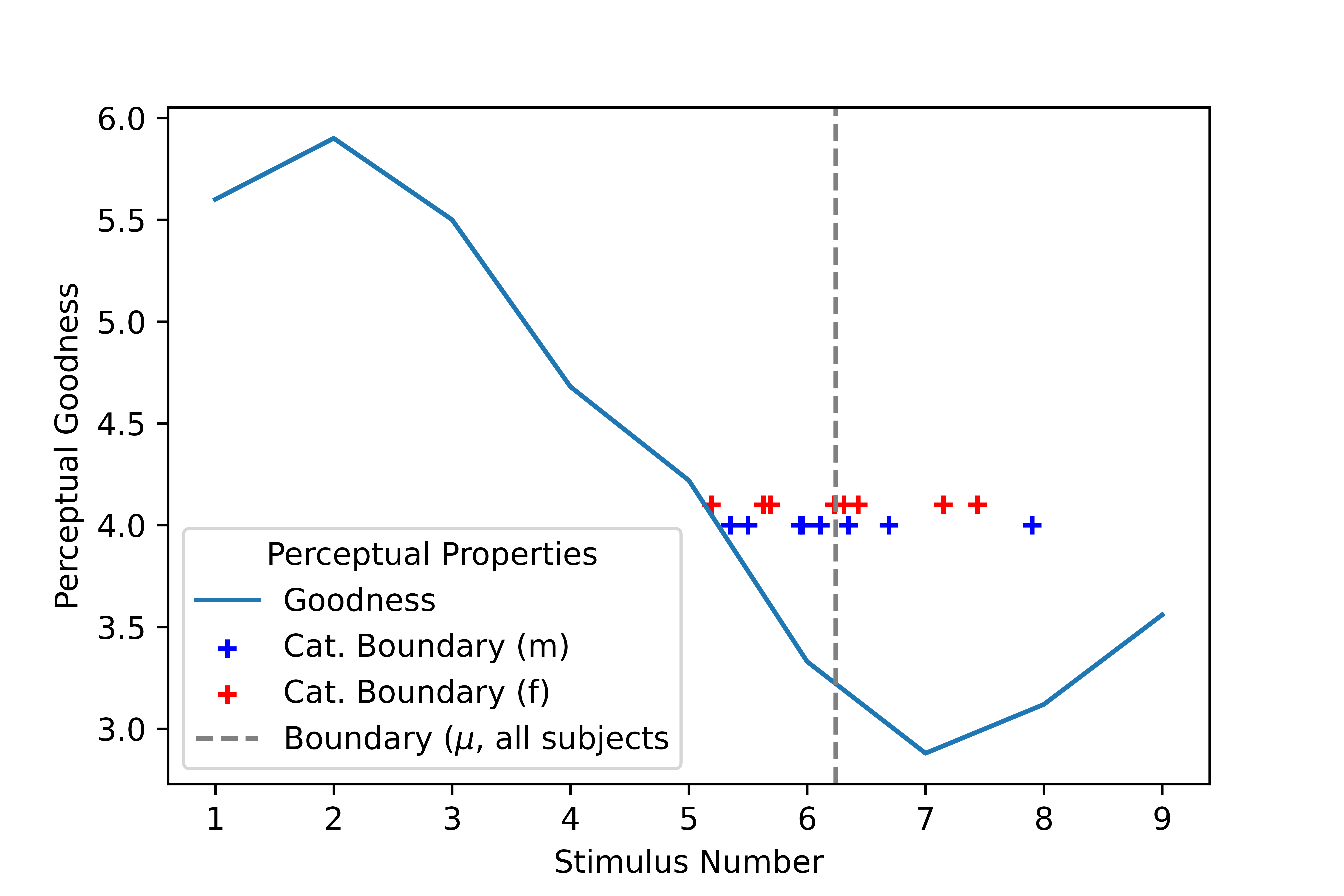}
    \caption{Goodness Ratings and Perceptual Boundary Locations}
    \label{Fig: Exp1 Perc}
    \end{centering}
\end{figure}

Stimulus perceptual goodness ratings were analyzed to understand the properties of the stimuli. Subjects rated the quality of the stimuli differently: perceived goodness is shown in Fig. \ref{Fig: Exp1 Perc} as the solid line, which is highest for stimuli near the /i/ prototype, and lowest for stimuli falling closer to /\textipa{I}/ in perceptual space. Individual differences were observed in the location of the boundary between /i/ and /\textipa{I}/.\footnote{To locate each individual’s /i-I/ boundary, a probit function was estimated for each subject to find the stimulus number which yielded a 50\% probability of categorization of the stimulus as /\textipa{I}/: \begin{equation}\Phi(/\textipa{I}/|stim_no) = \alpha + \beta * stim_no\end{equation}.} All subjects’ boundaries fell between stimuli 5 and 8, shown with blue (males) and red (females) crosses in Fig. \ref{Fig: Exp1 Perc}. No obvious gender differences in category location were found, so we use the grand mean to represent the perceptual boundary (grey dashed line) in our further analysis.  

Natural vowel productions and mimicry tokens were digitized at 9600 Hz and acoustically analyzed. Frequencies for the first two formants of natural productions and mimicry productions were then measured from wide-band spectrograms produced using Xwaves running on a Sun Ultra10 workstation, with a 4ms Hamming window for female subjects and an 8ms window for males. Formant frequencies were measured at a point located at the 10th vocal period after the onset of voicing in the first formant. 

\subsubsection{Results}\label{Sec: Exp1 Results}
\begin{figure}
    \begin{centering}
    \includegraphics[width=\columnwidth]{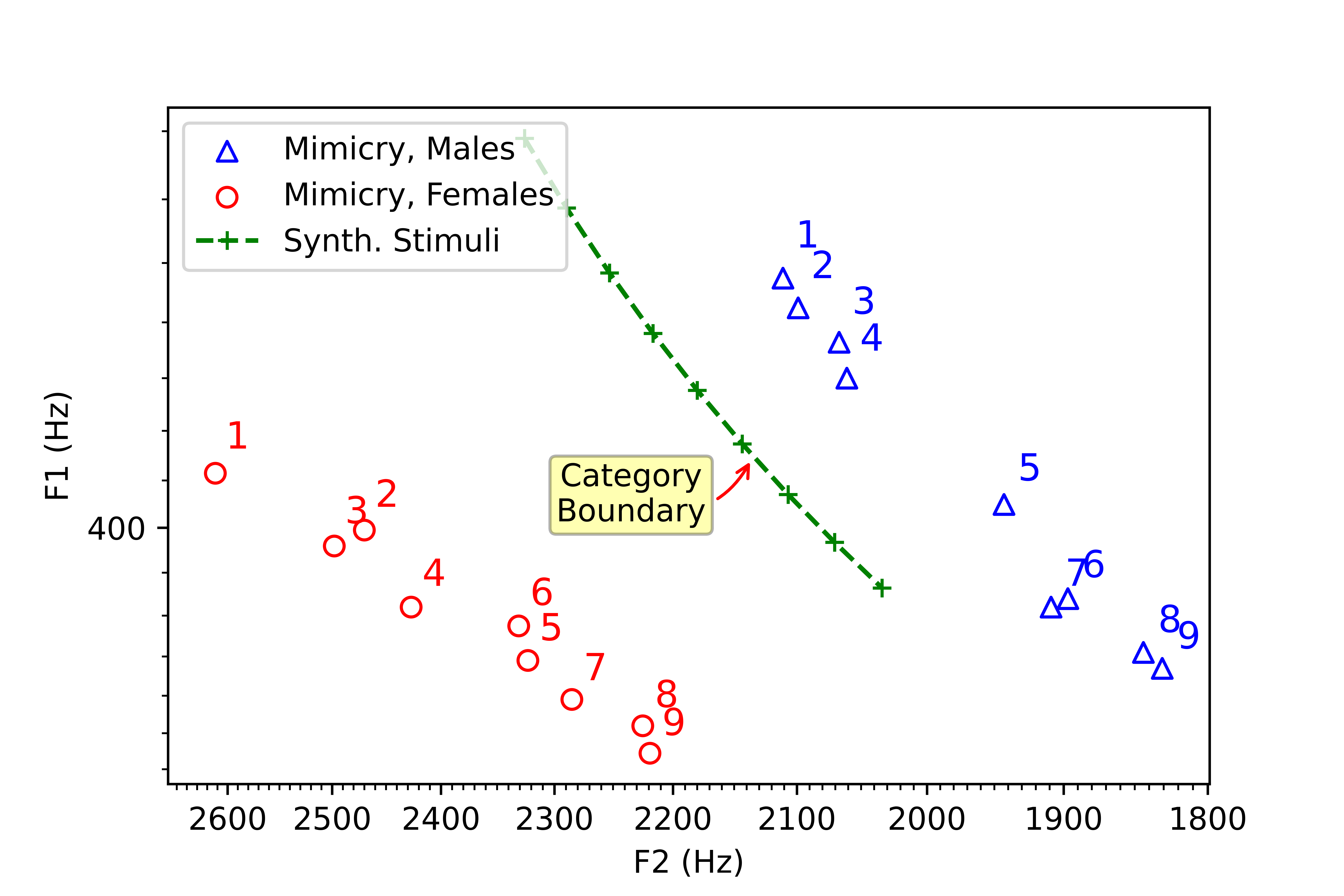}
    \caption{Mimicry Responses}
    \label{Fig: Exp1 Mimicry}    
    \end{centering}
\end{figure}

Mean formant frequency (Hz) of mimicry responses to the stimulus series for males and females are shown in Fig. \ref{Fig: Exp1 Mimicry}.  The stimulus series is shown with green crosses. Numbered red circles show mean mimicry responses for all female subjects to each stimulus, and numbered blue triangles show responses for males. Because the responses are in correct numerical order (with the exception of responses to stimulus 6 for females) it is readily apparent that subjects used sub-phonemic control of productions when mimicking. The greater separation between responses to stimuli 1-4 vs. 5-9 for males suggests that males weighted categorial properties of the stimuli more heavily than females, whose mimicry outputs appear more continuous. 

From the acoustic properties of the mimicry productions, Just Producible Difference (JPD) limens were next calculated as follows. Using the methodology of Flanagan \cite{flanagan1955difference}, pair-wise within-subject comparison was performed of each response vowel to every other. (The first production in each pair is referred to as the reference response, and the second is the comparison response.) For example, for subject 1, first each mimicry reference response to stimulus 1 was compared to each comparison response to stimuli 2-9, and so on for all reference stimuli for all subjects. The comparison response was deemed different from the reference response if it varied by more than a threshold: 81.3 Hz in F1, or 161.4 Hz in F2.\footnote{The thresholds were set by summing expected speaker variability and measurement error. Speaker variability in formant response within repeated productions of the same sound by the same speaker in the same context has been reported as F1: 40Hz and F2: 140 Hz by \cite{broad1976toward} Measurement error estimation is described in Fn\footref{fn_error}}

Pair-wise within-subject difference data were grouped according to the identity of the reference and comparison stimuli. The mean of each group indicates the probability of producing a response to the comparison stimulus which is different from the reference response, and the distance between the reference and comparison stimulus in mels is noted.

\begin{figure}
    \begin{centering}
    \includegraphics[width=\columnwidth]{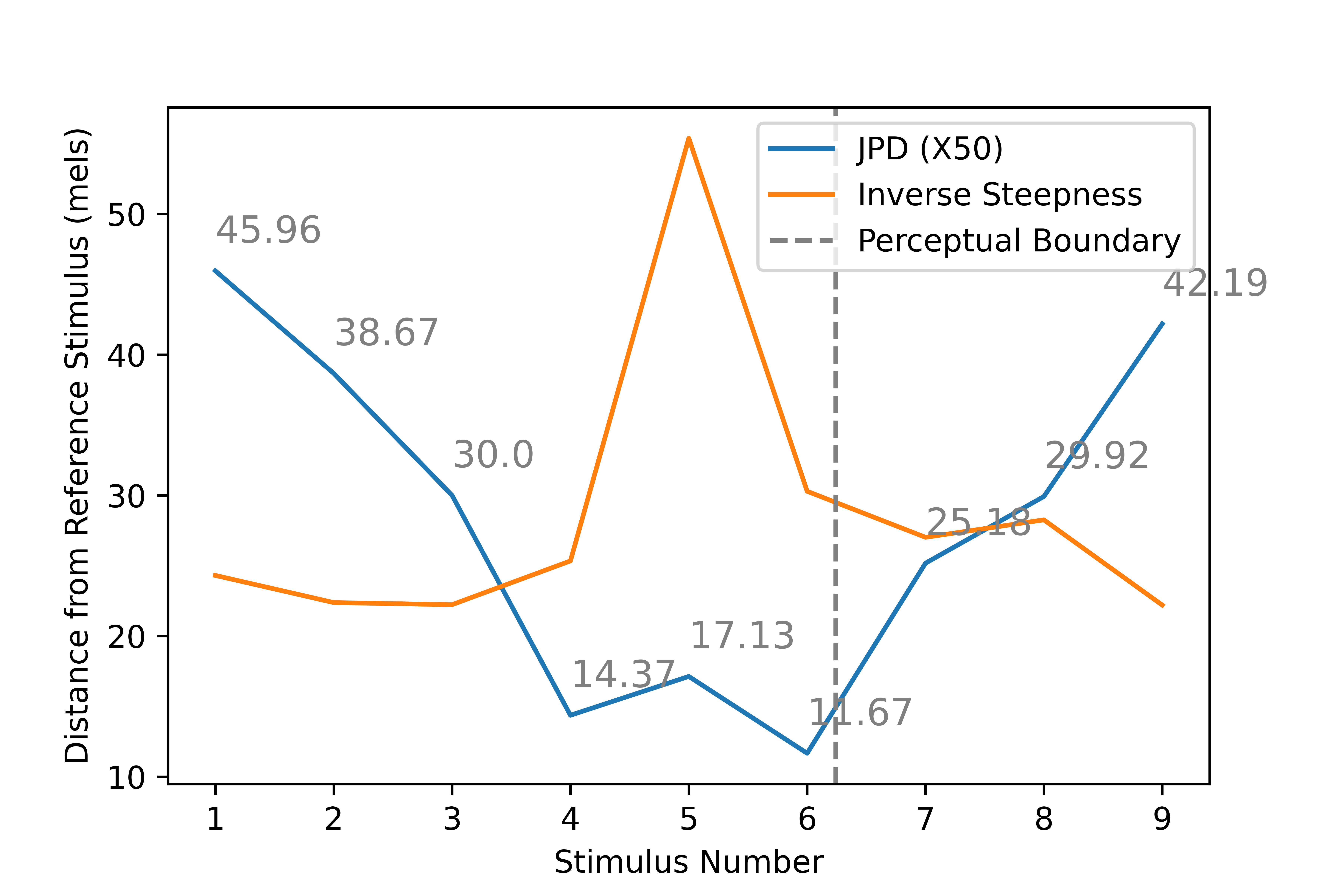}
    \caption{Just Producible Difference Limens Psychometric Function Inverse Steepness}
    \label{Fig: Exp1 JPD}
    \end{centering}
\end{figure}

Probit models for each reference stimulus were next estimated from the reference-comparison probabilities of difference in order to characterize them as a psychometric function \cite{levitt_1971} and so obtain numerical estimates of JPD. The location of the difference limen is found by calculating the mel distance at each point along the stimulus line which yields 50\% probability of different responses ($X^{50}$), and the steepness of the difference curve varies inversely with the distance between $X^{75} - X^{50}$. Since the lowest mean probability of difference over the entire stimulus series was observed as 10\% for repeated imitations of the same stimulus in Exp 1 and 2, it was assumed that a floor effect obtained due to natural variability in sequential productions of a sound, and a floor term $c=.1$ was included in the probit function used to estimate the difference limens (Eq \ref{Eq: JPD Probit}).
\begin{equation}\label{Eq: JPD Probit}\Phi(diff | Ref Stim, Comp Stim) = c + (1 -c)(\alpha + \beta * d_{rc})\end{equation}
($d_{rc}$ is the distance in mels between reference and comparison stimuli.) Separate probit models were estimated for each reference stimulus to derive JPD and inverse steepness.  Results for all subjects are shown in Fig. \ref{Fig: Exp1 JPD}.

\subsubsection{Discussion}\label{Section: exp1 discussion}
The JPD estimate varies between 45.96 mels (reference stimulus 1) and 11.67 mels (ref stim 6). There is a clear pattern of influence of perceptual category location and structure on mimicry outputs: subjects could better mimic differences in stimuli near the category boundary, shown by the smaller difference limen and correspondingly larger inverse steepness. Near the two stimulus series endpoints, subjects did not produce such fine-grained differential responses.  This effect resembles the better-studied perceptual magnet effect \cite{kuhl1991human}, perhaps reflecting underlying commonality between the perception and production systems. The size of JPD appears to differ from JND, however, by up to a factor of 5: observed JNDs for vowel formant frequencies vary by experimental procedure \cite{hawks_1994,hermansky_1987}, and are variously reported  as between 12-28 Hz for F1, and 20-90 Hz for F2 \cite{hawks_1994,kewley1999vowel}.

The JPD estimation procedure in Exp 1 was hampered by several factors.  First, male and females seemed to use different mimicry strategies, possibly since the stimuli sounded male-like. Furthermore, one end of the stimulus series falls in a region of the vowel space where quantal and saturation effects limit the amount of differentiation possible \cite{STEVENS_1989}; the other end was a vowel which does not normally appear in isolation in AE.   Experiment 2 was designed to overcome these limitations. 

\subsection{Experiment 2}
Experiment 2 used the same mimicry paradigm as in Exp 1, but with stimuli that were improved to overcome weaknesses identified in \S\ref{Section: exp1 discussion} above, with the goal of clarifying the JPD estimate. Custom stimuli were created for each subject resynthesized from their natural productions, whose endpoints were /\textipa{I}/ and /\textipa{E}/ enclosed in CVC word frames, extending past the category prototype on both ends. It was hoped these improvements would give a clearer measurement of the psychometric JPD function. 

\subsubsection{Stimuli}\label{sec: exp2_stimuli}
A custom, fourteen-step synthetic continuum, with vowels embedded in closed syllables varying between [\textipa{hId}] and [\textipa{hEd}], and extending slightly past those word prototypes on either side, was prepared for each subject.  The stimuli differed only in the F1 and F2 frequencies of the fricative and vowel portion of the utterance. From recordings of each subject’s natural productions of the words ‘hid’ and ‘head’, mean formant frequencies were calculated. The distance in F1 and F2 (Hz) between the mean productions was divided by 10 to yield the step sizes for the continuum. The most auditorily robust production of ‘hid’ was chosen as the base for the continuum.\footnote{Robustness criteria were lack of clipping of the waveform, longer duration, and clearest F1-F3 frequencies.}

Start- and end-points for the /hI/ portion of this base token were determined visually from wide-band spectrograms.  This base token was then resynthesized fourteen times, varying the frequencies of F1 and F2 each time in increments of the F1 and F2 step size. Stimulus 1 was resynthesized with no change in formant frequencies, and thus closely approximated the subject’s natural production of ‘hid’. Stimuli 2 through 12 were resynthesized by increasing F1 by the F1 step size and decreasing F2 by the F2 step size. Stimulus 10 thus approximates the subject’s natural production of ‘head’, while stimuli 11 and 12 are somewhat closer to ‘had’. Stimuli 0 and -1 were resynthesized by decreasing F1 and increasing F2, and were thus closer to the subject’s production of ‘heed’.

All stimuli were resynthesized using Praat \cite{boersma_praat} using the following steps. The base token was resampled at 11 kHz, and its formants were extracted using the Burg algorithm with a 25ms Gaussian window and a time-step of 10ms. The formant contour was used to create a filter. The source characteristics of the sound were extracted using inverse filtering.  The LPC coefficients for the token were also extracted using the Burg algorithm. The token then could be resynthesized from the source characteristics, the LPC coefficients, and the formant filter. Modified tokens were created by modulating the frequencies of the formant filter.

Post-hoc analysis of the Exp 2 stimulus formant values showed that the resynthesis did not modify F2 as effectively for female subjects as for males.  Most of the variation in the stimulus series for females was captured in the F1 dimension. The cause of this problem is not clear but may reflect shortcomings in LPC coding for female speech using the Burg algorithm.

\subsubsection{Procedure}
Recording Sessions. On Day 1, subjects recorded in random order six tokens each of utterances containing the eleven monophthongs of AE in hVd frames, yielding a total of 66 natural production tokens per subject.  The sentence frame for each utterance was the same, ‘Say the word hVd again.’ 

Perceptual Testing. On Day 2, subjects categorized each synthetic token from their own custom stimulus series, and rated the word’s goodness on a 3-point scale. Tokens were presented to subjects six times each in random order. Subjects were instructed to choose which word best represented the sound they had just heard.\footnote{The eleven choices were labeled ‘heed’, ‘hid’, ‘head’, ‘hayed’, ‘had’, ‘hod’, ‘hawed’, ‘hoed’, ‘hood’, ‘who’d’, ‘HUD’.} In ambiguous cases, subjects were instructed to guess. 

Mimicry. After perceptual testing on Day 2, subjects mimicked the custom stimulus series with their productions recorded for further study. Each stimulus was presented to the subjects six times in random order, for a total of 84 imitations per subject.  The subject pressed a button on the screen indicating readiness and the stimulus was played over the headphones.  The subject was instructed to imitate it back into the microphone without delay. The productions were digitized directly using the workstation’s A/D converter.

\subsubsection{Analysis}
\begin{figure}
    \begin{centering}
    \includegraphics[width=\columnwidth]{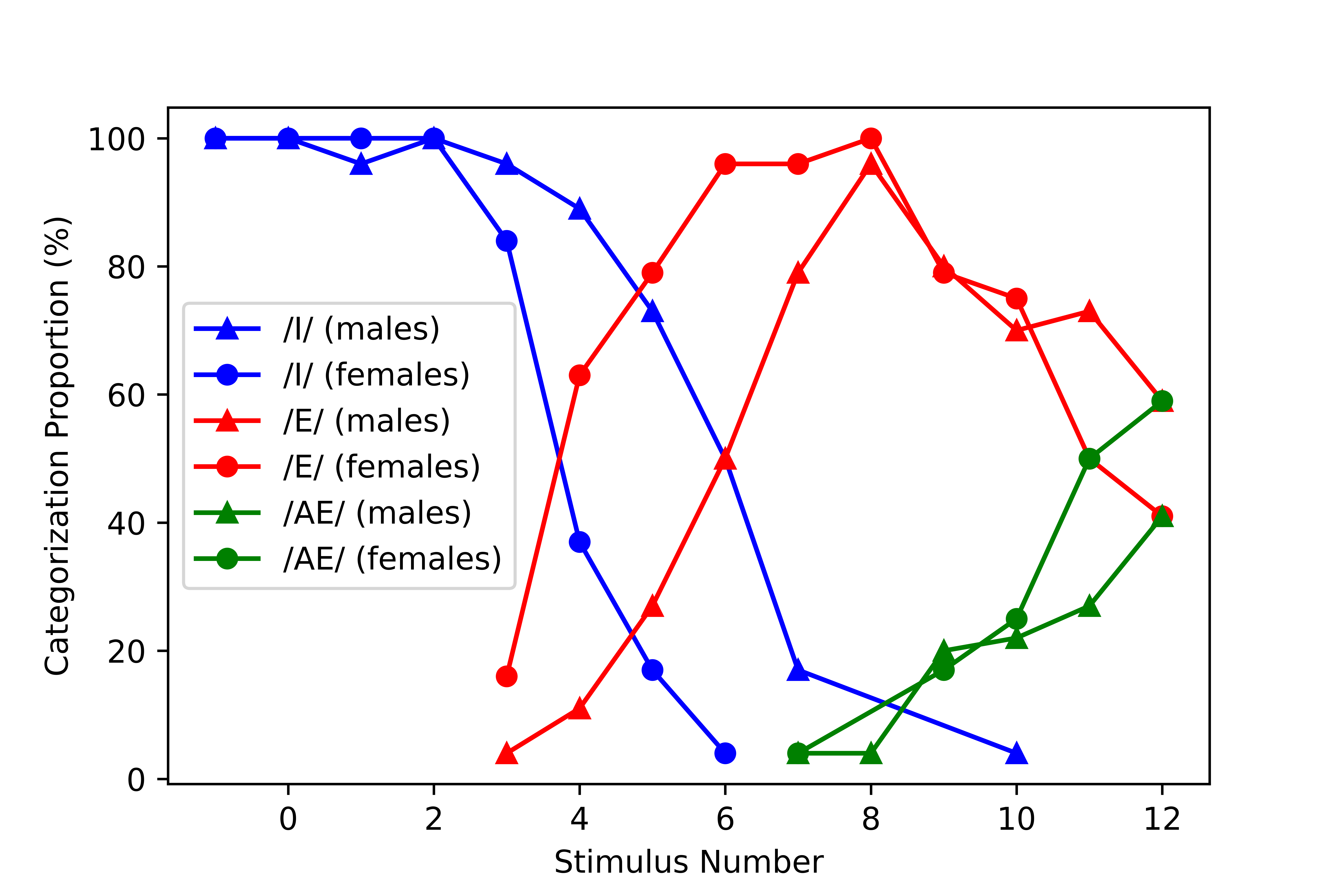}
    \caption{Experiment 2 Perceptual Categorization of Custom Stimuli}
    \label{Fig: Exp2 Categorization}
    \end{centering}
\end{figure}

Fig. \ref{Fig: Exp2 Categorization} presents the categorization functions for male (triangles) and female subjects (circles). As in Experiment 1, the stimuli were uniformly perceived as speech-like, and all subjects were able to categorize the stimuli in the forced-choice paradigm as expected, and preferred /\textipa{I}/, /\textipa{E}/, or /\ae/ in almost all cases. Though an attempt was made to create /\textipa{i}/-like stimuli at the top end of the series, only a negligible quantity of tokens were classed as /\textipa{i}/. The perceptual prototype for /\textipa{E}/ occurred earlier in the stimulus series than expected (near Stimuli 8 and 9, vice 10). The stimulus series did not extend fully into the center of the /\ae/ target for subjects, and the probability of categorization stimuli as /\ae/ never exceeded 60\%. 

There is a significant difference in the location of the perceptual boundary between /\textipa{I}/ and /\textipa{E}/ for males and females, with the male boundary located near stimulus 6, and the female boundary located near stimulus 4. A possible source of this difference is the lack of successful modulation of F2 frequency in the resynthesized female stimuli (see \S\ref{sec: exp2_stimuli}).  

Perceptual goodness ratings were analyzed for all subjects.  Males and females showed equivalent goodness ratings for the whole stimulus series, unlike in Experiment 1 where one end of the stimulus series was rated worse by females than males.\footnote{To confirm this observation, ANOVA of goodness rating as a function of subject gender, vowel, and the interaction of $sex \times vowel$ was performed.  The model as a whole was significant ($F=15.3, Pr>F .0001$), as were the main effects of gender ($F=17.23, PR > F .0001$) and vowel ($F=27.09, PR > F .0001$), but the interaction of $gender \times vowel$ was not significant ($F=1.57, PR > F .2162).$}

Mimicry responses were then acoustically analyzed, and their first two formant frequencies extracted, using the procedure described in \ref{Sec: Exp1 Analysis}.\footnote{Formant frequency measurement error was estimated by independently remeasuring mimicry productions from Exp2 female speakers.  The mean F1 and F2 frequencies for these vowels when re-measured were within 10\% of the standard deviation.\label{fn_error}} 

\subsubsection{Results}

\begin{figure}%
\centering
\subfigure[Males]{%
\label{Fig: Exp2 Mim Males}%
\includegraphics[width=0.4\textwidth]{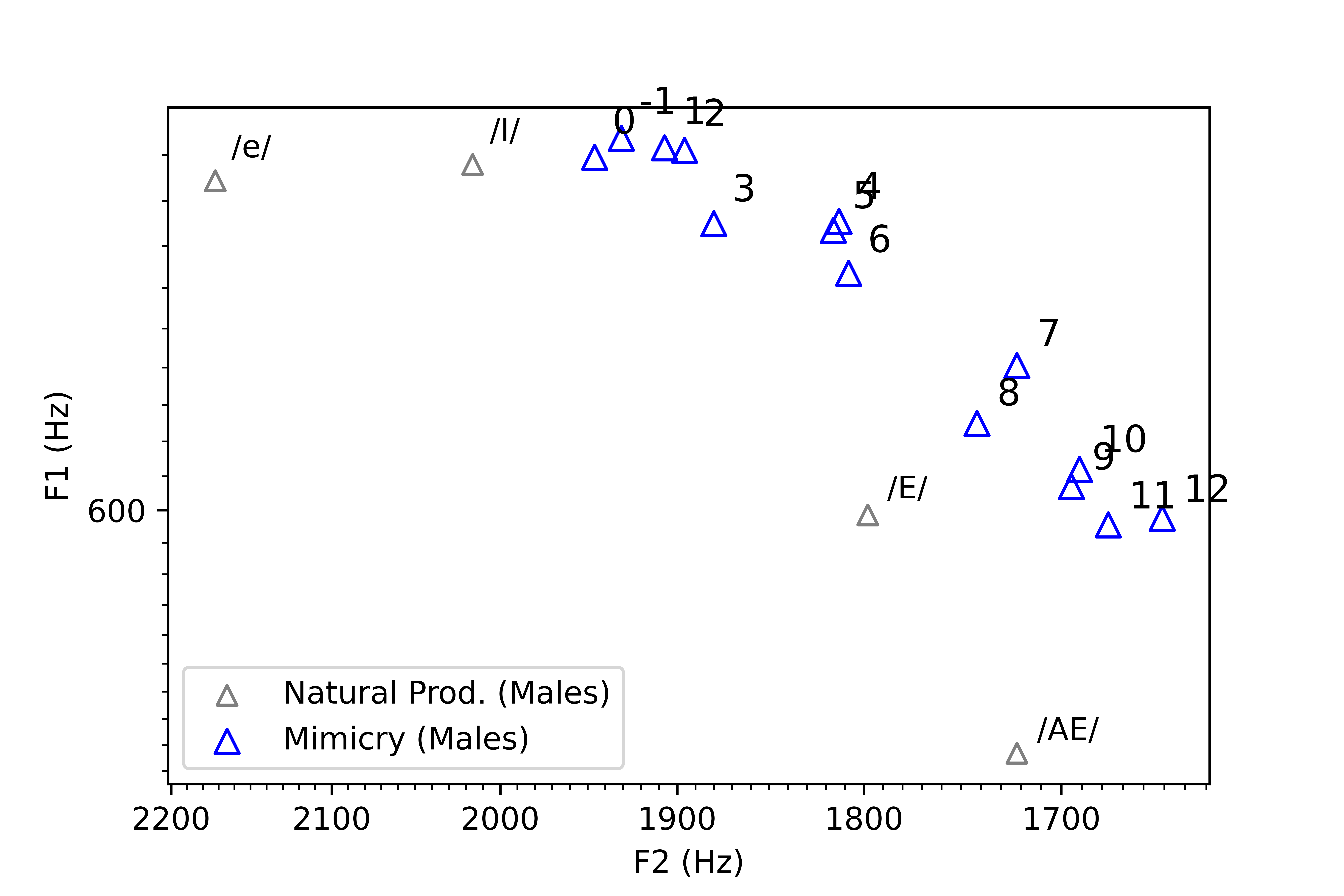}}%
\qquad
\subfigure[Females]{%
\label{Fig: Exp2 Mim Femles}%
\includegraphics[width=0.4\textwidth]{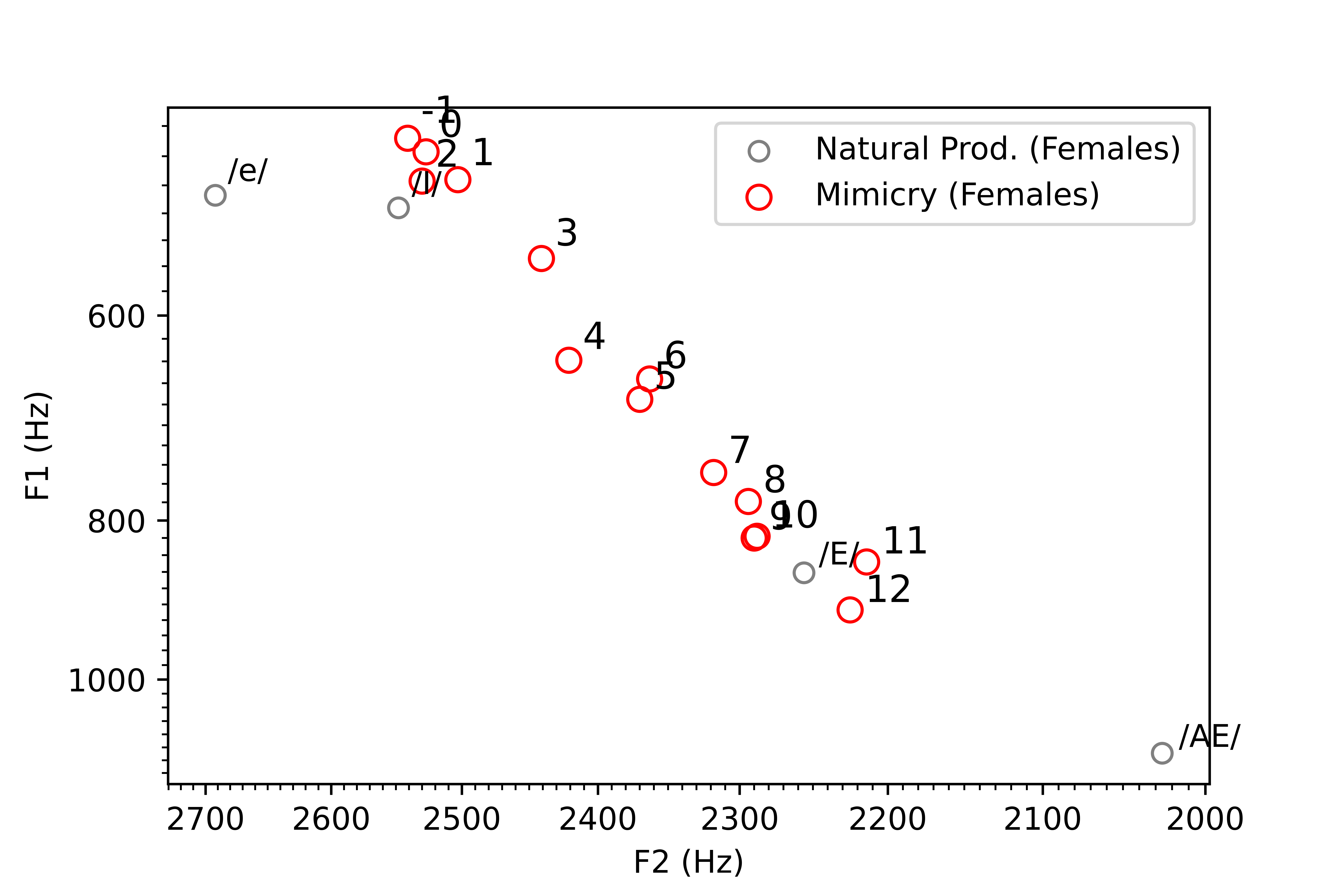}}%
\caption{Natural Production and Mimicry Responses}
\end{figure}

Male mimicry responses are shown in Fig. \ref{Fig: Exp2 Mim Males} and female responses are shown in Fig. \ref{Fig: Exp2 Mim Femles}.  

As in Exp 1, the mimicry responses show influence of the phonemic categories /\textipa{I}/ and /\textipa{E}/, as well as sub-phonemic control leading to approximate linear ordering of response means. Some asymmetry is apparent in the placement of the mimicry responses relative to natural productions for males and females, with female responses to lower-numbered stimuli located higher in the vowel space. This asymmetry is likely explicable from the differences in location of the phonemic category boundary (see Fig. \ref{Fig: Exp2 Categorization}). 

\begin{figure}
    \begin{centering}
    \includegraphics[width=\columnwidth]{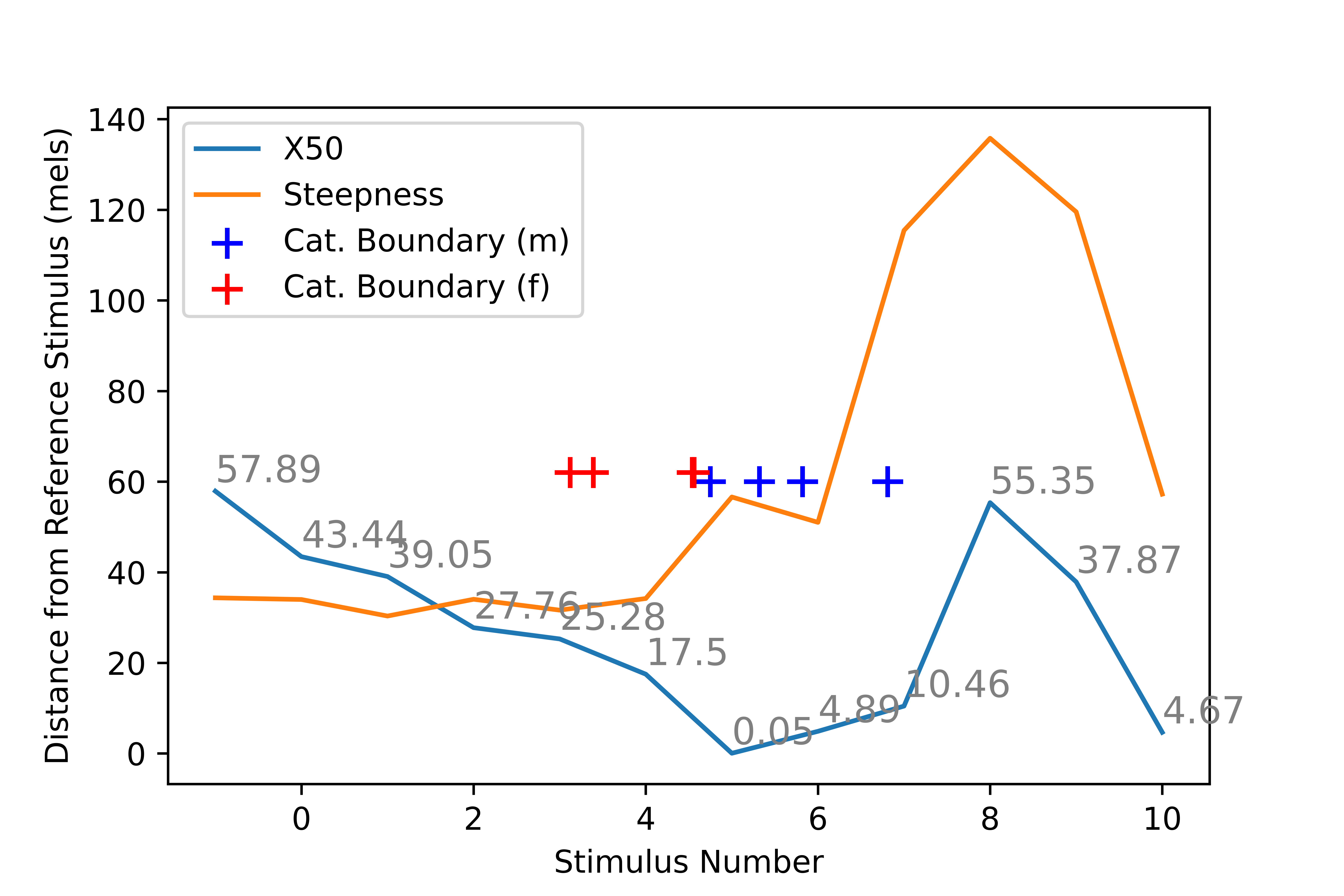}
    \caption{Experiment 2 Just Producible Difference and Inverse Steepness}
    \label{Fig: Exp2 JPD}
    \end{centering}
\end{figure}

Probability of difference data from pairwise comparisons of reference and comparison responses were tabulated using the procedure described in \ref{Sec: Exp1 Results} above. As in Exp 1, probability of difference was lower near the category prototypes and higher near the category boundary.  The probability surface was used to estimate the difference limen as a psychometric function, using the same functional form equation \ref{Eq: JPD Probit}. Location of $X^{50}$ and inverse steepness ($X^{75} - X^{50}$) is shown in Fig. \ref{Fig: Exp2 JPD}.\footnote{Estimates of \ref{Eq: JPD Probit} did not converge for stimuli 11 and 12, so are not reported.}

$X^{50}$ location shows the same pattern as in Exp 1, with peaks near the phonemic prototypes and valleys near the phonemic boundaries.  However the inverse steepness only follows the expected pattern for the /\textipa{I}/-like portion of the stimulus series, stimuli -1 to 4 (Fig. \ref{Fig: Exp2 JPD}).

\subsubsection{Discussion}
Some aspects of the Experiment 2 procedure indeed worked better than Experiment 1.  The stimulus continuum was more comprehensive, extending past the vowel prototype center on either end, unlike in Experiment 1 where the stimulus series was limited by the saturation ceiling around /\textipa{i}/. Furthermore, the stimuli were not rated differentially for quality by male versus female subjects.

However, several other aspects of the Experiment 2 stimulus creation procedure were less successful and did not mitigate issues identified in Experiment 1. The stimulus series here contained 2 complete and 1 partial third phonemic target, /\textipa{I}/, /\textipa{E}/, and /\ae/. But the LPC resynthesis technique used to create the custom stimuli for each subject did not effectively modify F2 for females. Furthermore the functional form chosen for estimating the difference limens is unable to accommodate a function with more than one step.  

Because the JPD estimation procedure was not successful for the /\textipa{E}/- and /\ae/-like portion of the stimulus series, we view JPD estimates from Exp 2 as tentative and only report upper and lower bounds for Exp 2 stimuli in the /\textipa{I}/ region of Experiment 2. We leave improvements to stimulus creation and JPD estimation for future research.

\section{Conclusion}
The JPD estimates from Exp 1 and from the /\textipa{I}/-like stimuli in Exp 2 are similar in magnitude and vary according to their location in the phonemic/perceptual space. Our combined estimates of the granularity of the vowel production space are shown in Table \ref{Tab: JPD}.

\begin{table}[!t]
\renewcommand{\arraystretch}{1.3}
\caption{Just Producible Difference (JPD) Limens}
\label{Tab: JPD}
\centering
\begin{tabular}{|c||c|c|c|}
\hline
\thead{JPD Limen (mels)} & \thead{Exp 1} & \thead{Exp 2} & \thead{Mean}\\
\hline
Upper Bound	& 45.96 & 57.89 & 51.93\\
\hline
Lower Bound & 11.67 & 17.50 & 14.59\\
\hline
\end{tabular}
\end{table}

Taken together show that the speech production system is up to five times less accurate than the perceptual system in distinguishing between vowel stimuli.  Fidelity of the transfer function between auditory input and production output (Fig \ref{Fig: Mimicry Transfer Function}) is lowest near the perceptual prototype of the vowel category and highest near the perceptual boundary.

This finding has implications for episodic theories of speech production. Hybrid exemplar models as delineated by \cite{GOLDRICK2023101254}, in which both continuous and symbolic representations emerge, co-exist, and guide production, account well for the phonemic and sub-phonemic control evinced by the subjects in the current study. Within such models our results gives a first estimate of the level of granularity which is encoded by the non-symbolic portions of those representations. Future research could use mimicry to elucidate the processes underpinning exemplar storage.

A second implication of this finding is that it provides a theoretical explanation for widely observed properties of vowel systems studied cross-linguistically.  The tendency of stable vowel systems with more than eight primary vowels to recruit a third dimension of vowel color follows from the finding that the speech motor control system can only produce stable differences between vowel exemplars which are at least $\sim$50 mels apart. The seeming upper bound of 4 front vowels in most stable systems also follows from the intervowel distance of 50 mels.

A final implication is for diachronic patterns of vowel shifts and resulting mergers. Among the various factors that predict neutralization in vowel contrast \cite{wedel_2012} \textemdash notably functional load and lexical frequency \textemdash The estimate of JPD predicts that vowels closer than $\sim$50 mels are susceptible to merger \cite{lubowicz_2011}.

\section{Limitations}

The current findings have some obvious limitations in scope and procedures.  The scope of the current study is limited because of the vowels used and the subjects recruited. We only report results for mid and high front vowels for speakers of American English.  We would expect differences in the granularity of the vowel space in different regions, and following different axes (for example, high synthetic vowels varying between /i/ and /u/). We do not know \textit{a priori} how speakers of other languages \textemdash with potentially fewer phonemic distinctions of their vowel space \textemdash would perform on this task.

The current study is also limited by the measurement and estimation procedures used to observe JPD. Better observations of JPD would do some or all of the following: measure probability of difference approaching each reference stimulus from both sides; enforce a directionality constraint on phonetic difference, treating as different only a response which is on the correct side of the reference stimulus; use an adaptive testing procedure which modifies the interstimulus step size based on real-time analysis of each mimicry production \cite{hall_1981}.  We leave for future research these improvements which we believe would likely sharpen and strengthen the measurement of JPD.

\appendices

\section*{Acknowledgment}

The author would like to thank Gail Brendel Viechnicki for her support of this research; Ken de Jong, Yukari Hirata, Howard Nusbaum, and Hynek Hermansky for their encouragement; and Margaret Renwick for discussing early drafts of this work.

\ifCLASSOPTIONcaptionsoff
  \newpage
\fi

\newpage
\bibliographystyle{IEEEtran}
\bibliography{bibtex/bib/IEEEabrv,bibtex/bib/refs}

\begin{IEEEbiography}
[{\includegraphics[width=1in,height=1.25in,clip,keepaspectratio]{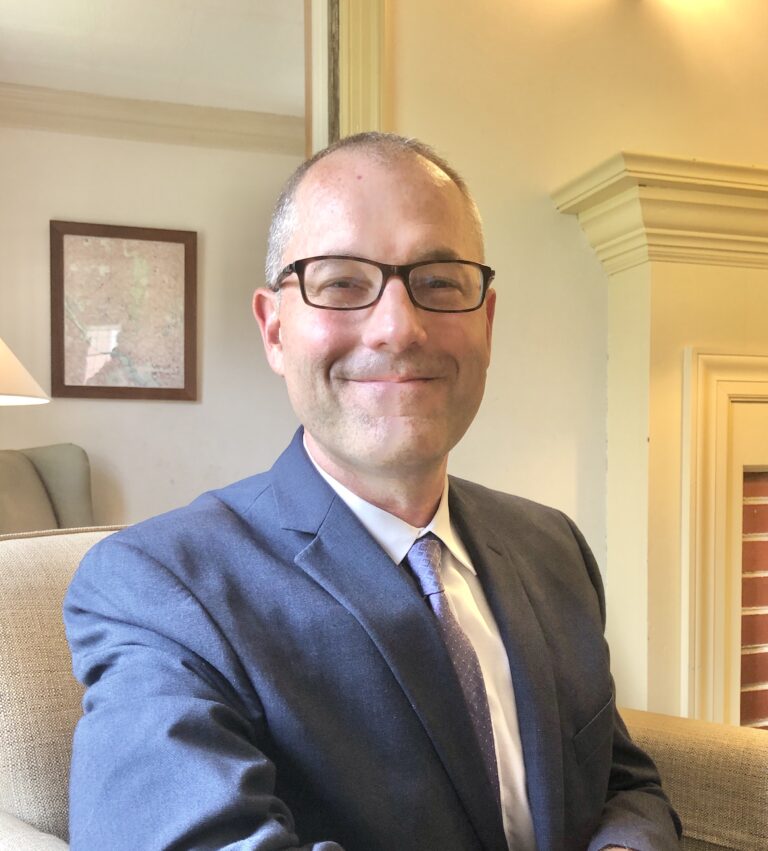}}]{Peter Viechnicki}
directs the Human Language Technology Center of Excellence in the Whiting School of Engineering at Johns Hopkins University in Baltimore, Maryland.
\end{IEEEbiography}

\end{document}